# Development of Fake News Model Using Machine Learning through Natural Language Processing

Sajjad Ahmed, Knut Hinkelmann, Flavio Corradini

***Abstract*—**Fake news detection research is still in the early stage as this is a relatively new phenomenon in the interest raised by society. Machine learning helps to solve complex problems and to build AI systems nowadays and especially in those cases where we have tacit knowledge or the knowledge that is not known. We used machine learning algorithms and for identification of fake news; we applied three classifiers; Passive Aggressive, Naïve Bayes, and Support Vector Machine. Simple classification is not completely correct in fake news detection because classification methods are not specialized for fake news. With the integration of machine learning and text-based processing, we can detect fake news and build classifiers that can classify the news data. Text classification mainly focuses on extracting various features of text and after that incorporating those features into classification. The big challenge in this area is the lack of an efficient way to differentiate between fake and non-fake due to the unavailability of corpora. We applied three different machine learning classifiers on two publicly available datasets. Experimental analysis based on the existing dataset indicates a very encouraging and improved performance.

***Keywords*—**Fake news detection, types of fake news, machine learning, natural language processing, classification techniques.

## I. Introduction

FAKE news is a type of yellow journalism or propaganda that consists of deliberate misinformation or hoaxes spread via traditional print and broadcast news media or online social media. Fake news is as old as the news industry itself—misinformation, propaganda, hoaxes and satire have long been in existence.

Today anybody can publish anything credible or not that can be consumed by the World Wide Web. Due to this, people can be deceived intentionally or unintentionally and do not think before sharing such types of news to the far ends of the world. The counterfeited news problem can be resolved or at least overcome with machine learning and artificial intelligence. In general, fake news detection is considered as a challenging task [26] that requires multidisciplinary efforts [25]. For deception detection, there exists a large body of research [32] done where machine learning methods are applied. Classification of online news and social media posts were the target of those methods but after the 2016 United States Presidential elections, determining fake news has also been the subject of attention in the literature. Simple content-related classification n-gram and part of speech (POS) tagging have proven insufficient in fake news context [1]. Fake news detection through classification is not sufficient since it missed the important context of the information, however a deep analysis of the content that can be useful [7]. Context-free grammar (CFG) produced good results with the combination of the n-gram in deception related classification. The accuracy achieved 85%-91% when applied on news article datasets through classification [2]. We propose a hypothesis that simple classification is not enough to tackle the issue; we need to combine it with machine learning techniques. The hypothesis is proven on publicly available datasets by developing the proposed model after several experiments.

We observe that the relative frequency of words can also be the reason for fake and non-fake class categorization. Using word cloud visualization, we observe the corpus trend, as shown in Fig. 1. The word cloud representation reflects important word entities. For example, we can easily observe the highly frequent words Political, Americas, 2016, President, Obama and Presidential Debates, respectively. We use different sources of news for the testing and training datasets so that we can observe how well our models generalize to unseen data points. In the first step, we applied text extraction features covered under the text classification module. Fake news can be categories in seven different types [3]. Table I explains seven types of fake news.

TABLE I
SEVEN TYPES OF FAKE NEWS [33]

| Sr. No | Type | Details |
|---|---|---|
| 1 | False Connection | When headlines, visuals or captions don't support the content. |
| 2 | *False Context | When genuine content is shared with false contextual information. |
| 3 | Manipulated Content | When genuine information or imagery is manipulated to deceive. |
| 4 | Satire | No intention to cause harms but has potential to fool. |
| 5 | *Misleading Content | Use of information to frame an issue. |
| 6 | Imposter Content | When genuine sources are impersonated. |
| 7 | Fabricated Content | New content that is 100% false, designed to deceive and do harm. |

*Types of news come under a political domain.

The rest of this paper is organized as follows, Second II reviews the previous work, and Section III describes the Methodology. The Proposed model, Preprocessing and Machine learning are described in Sections IV-VI, Section VII describes the implementation task, Results and discussion are described in Section VIII and finally, the last section gives the Conclusion and Future Work.

Sajjad Ahmed and Flavio Corradini are with the Department of Computer Science, University of Camerino, Camerino, 62032 Italy (e-mail: ahmed.sajjad@unicam.it, flavio.corradini@unicam.it).

Knut Hinkelmann is with FHNW University of Applied Sciences and Arts Northwestern Switzerland, 4600 Olten, Switzerland (e-mail: knut.hinkelmann@fhnw.ch).

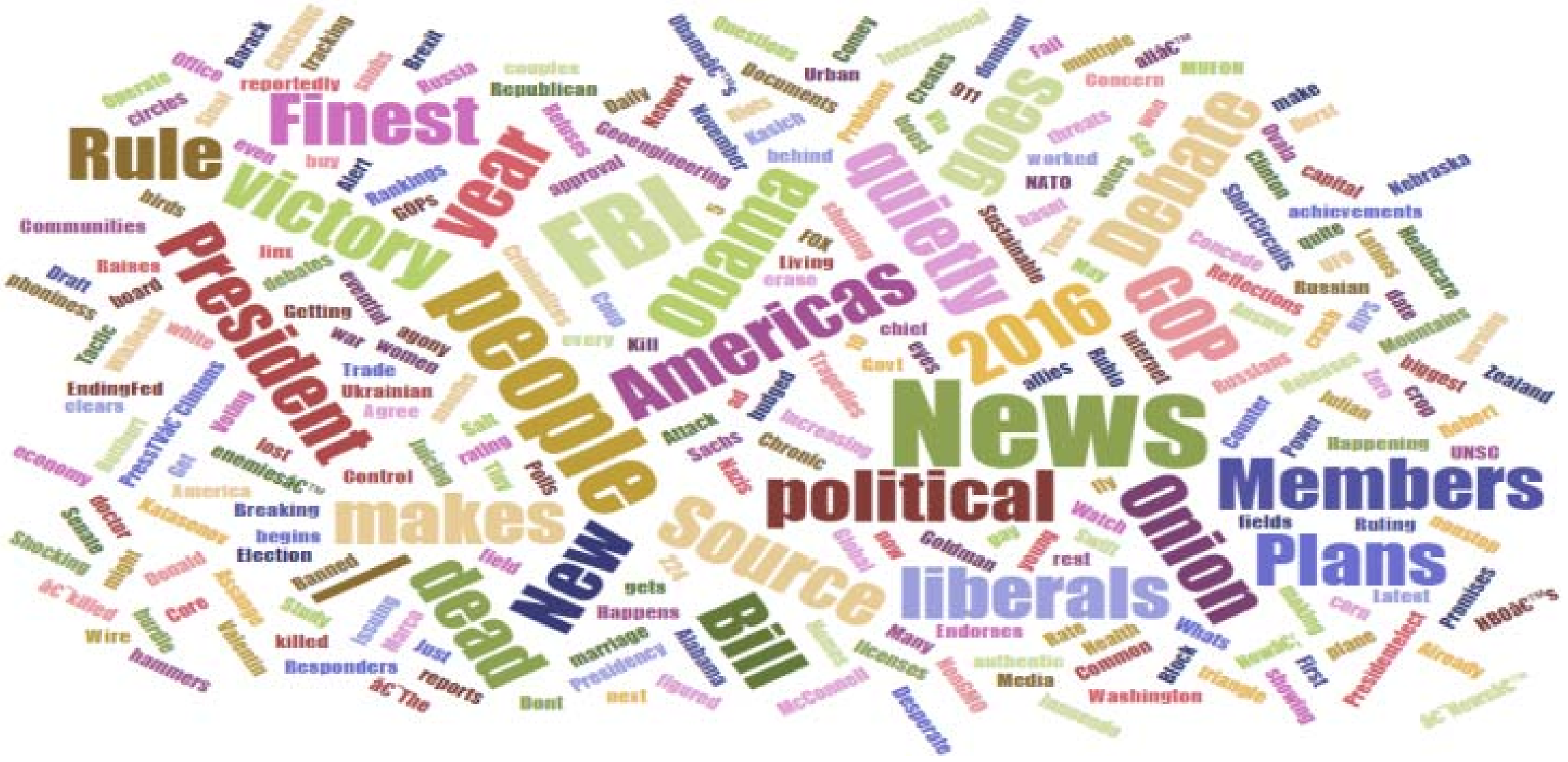

Fig. 1 Word Cloud of News Articles

## II. Literature Review

The problem addressed is very relevant in this information age, several previous works have been carried out from different perspectives, focused on different ways and using various techniques, but ultimately all seek to combat misinformation; some of these studies will be presented below. Traditional approaches based on verification by humans and expert journalists do not scale the volume of the news content that is generated online [4]. Text classification is the fundamental task in Natural Language Processing (NLP) and researchers have addressed this problem quite extensively [7]. Researchers proposed a model that can check the real-time credibility within 35 seconds after combining user-based, propagation-based, and content-based text [10].

The basic idea of Naïve Bayes is that all features are independent of each other [5]. Naïve Bayes needs a smaller data set and less storage space. Facebook post prediction through real or fake labeling can be done through naïve Bayes and it performs well [6]. A proposed method can separate fake contents in three categories: serious fabrication, large scale hoaxes and humorous fake [11]. It can also provide a way to filter, vet and verify the news. PHEME was a three-year research project funded by the European Commission from 2014-2017, studying NLP techniques for dealing rumor detection, stance detection [8] and [9], contradiction detection and analysis of social media rumors.

Fake news stories can be easily shared on social media platforms but it is difficult to identify fake content automatically. Using information sources (Visual cues & Cognitive cues) and social judgment (Cognitive, Behavioral & Affective), Facebook examines that machine learning classifiers can be helpful to detect fake news [13]. We preferred Support Vector Machine for fake news detection as it is a more researched algorithm nowadays. It is difficult to say that it is the best classifier in fake news because the selection of classifiers totally depends on the organizational requirements [14]. Stance detection of the headline for binary classification through n-gram matching can also be assessed after comparing “related” vs. “unrelated” pairs. This approach can be applied in the detection of fake news, especially clickbait detection. They used a dataset released by the organization Fake News Challenge (FNC1) on stance detection for experiments. The dataset is publicly available and can be downloaded from the corresponding GitHub page along with baseline implementation [15]. Deep learning using NLP for the detection of fake news and applied different models are presented, an assessment is made of which may be the option to obtain good results [16].

## III. Methodology

### A. Data Exploration

The dataset used for classification was drawn from a public domain. Fake news articles were collected from an open source Kaggle dataset [33] that was published during the 2016 election cycle. The collection is made up of 18000 news articles highlighted in Fig. 2 (a). These articles were collected from news organizations NYT, Guardian, and Bloomberg during the election period. Articles are separated through binary labels 0 and 1. The dataset is already sorted qualitatively with fake, non-fake and not clear labels. This division can be seen in Fig. 2 (a) where we have 15,115 articles from the false category and 1,846 from the true category. The remaining articles are classified as not clear due to some other reasons like unique ID missing, source not clear etc. The task itself leads to a quite imbalanced dataset, as shown in Fig. 2, wherefrom the total articles, roughly 12% are in the true category. This imbalance is typical in this task, and also seen in previous similar works [29], [30]. The second dataset contains 5000 real news articles collected from the Signal Media News dataset [25], in which 2,541 belong to the false class and 299 to the true class, as shown in Fig. 2 (b). We skipped the unclear class due to the missing values. Articles were collected from major news media organizations e.g. the Guardian, Bloomberg, New York Times, NPR, etc. The dataset was published in 2016 before and after the United States presidential elections.

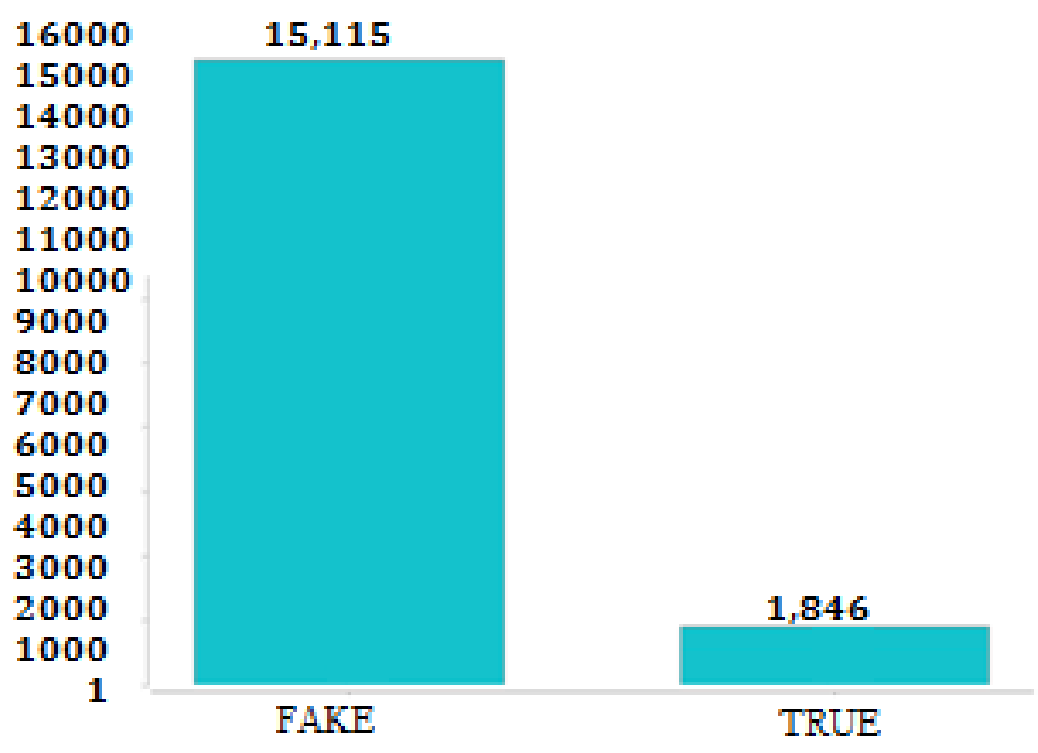

Fig. 2 (a) Class Distribution Kaggle Dataset (2016)

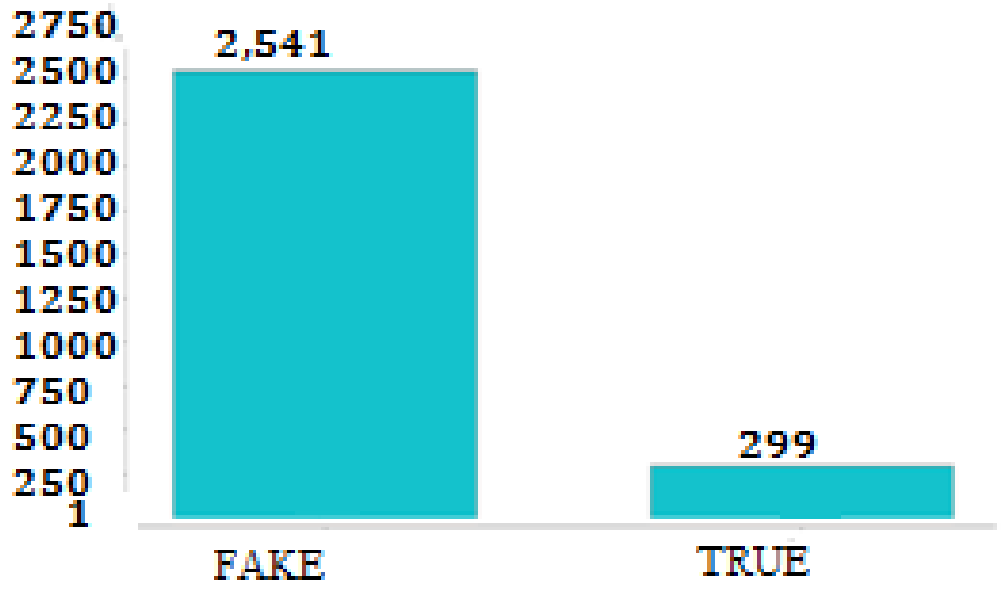

Fig. 2 (b) Signal Media News Dataset 2016

We used RapidMiner, a powerful machine learning tool for data exploration, preparation, information extraction, result visualization and result optimization. We analyzed the fake and true sentences through RapidMiner and initial results can be seen in Fig. 3.

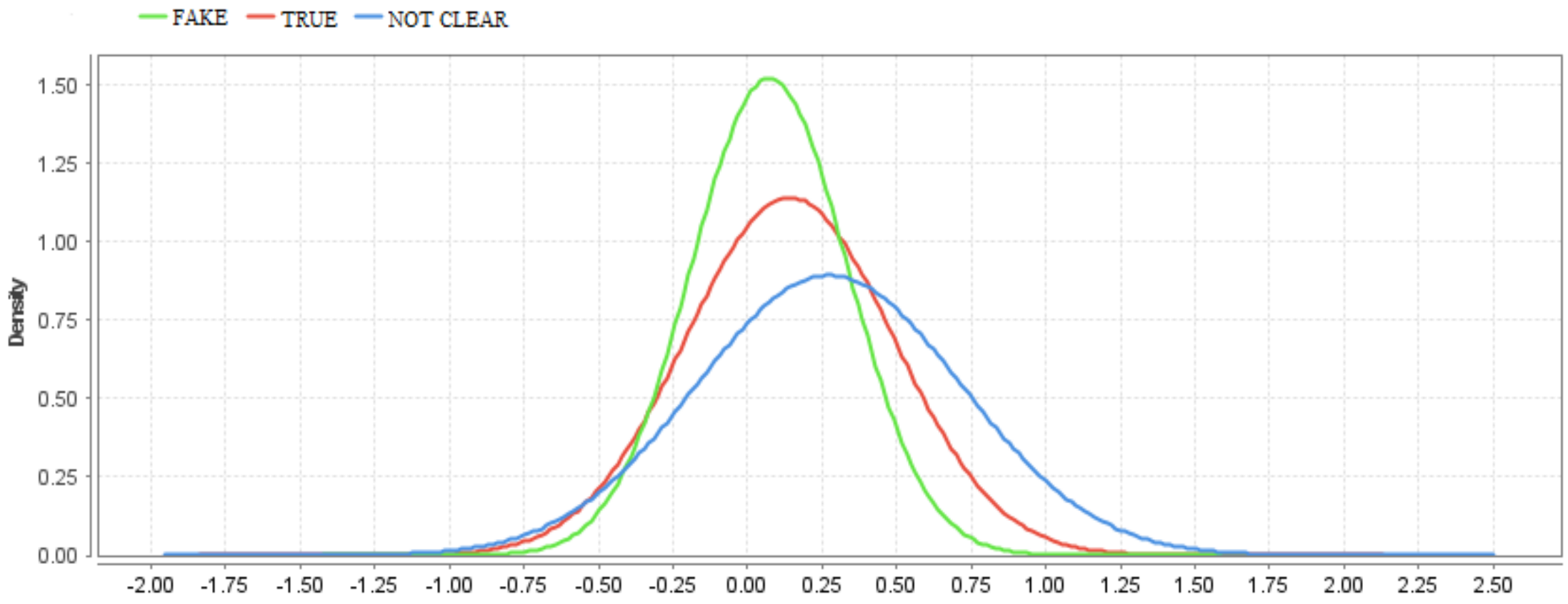

Fig. 3 Dataset class labeling chart

*B. Models Description*

Different classification models can be applied in this case, but to choose the most adequate one and to tune its parameters we run several experiments on different models. We started experimenting with classification models that have proven to be effective and give good results in related sentence classification tasks. Some of the models did not give good results and were discarded, one of them was Logistics Regression, while Support Vector Machines, naïve Bayes and Passive Aggressive gave promising results and we continued to experiment on them. To check the accuracy, we compare our results with other datasets through performance metrics.

- Naïve Bayes: It is a powerful classification model that performs well when we have a small dataset and it requires less storage space. It does not produce good results if words are co related between each other [18]. Fig. 5 contains the Naïve Bayes formula that explains the probability of an attribute that belongs to a class independent from other classes.
- Support Vector Machine: It performs supervised learning on data for regression and classification. The SVM computes the data and converts it into different categories. The advantages of Support Vector Machine are learning speed, accuracy, classification and tolerance to irrelevant features [19]. Support Vector Machine is one of the most researched classifiers nowadays and it performs well in the fake news detection problem [20].

Likelihood — Class Prior Probability

$$P(c \mid x) = \frac{P(x \mid c)P(c)}{P(x)}$$

Posterior Probability — Predictor Prior Probability

$$P(c \mid X) = P(x_1 \mid c) \times P(x_2 \mid c) \times \cdots \times P(x_n \mid c) \times P(c)$$

Fig. 4 Naïve Bayes formula

- Passive Aggressive: These algorithms are mainly used for classification [27]. The idea is very simple and the performance has been proven with many other alternative methods like Online Perceptron and MIRA.
- Logistic Regression: It is used to estimate the relationship between variables after using statistical methods. It

performs well in binary classification problems because it deals with classes and requires a large sample size for initial classification.

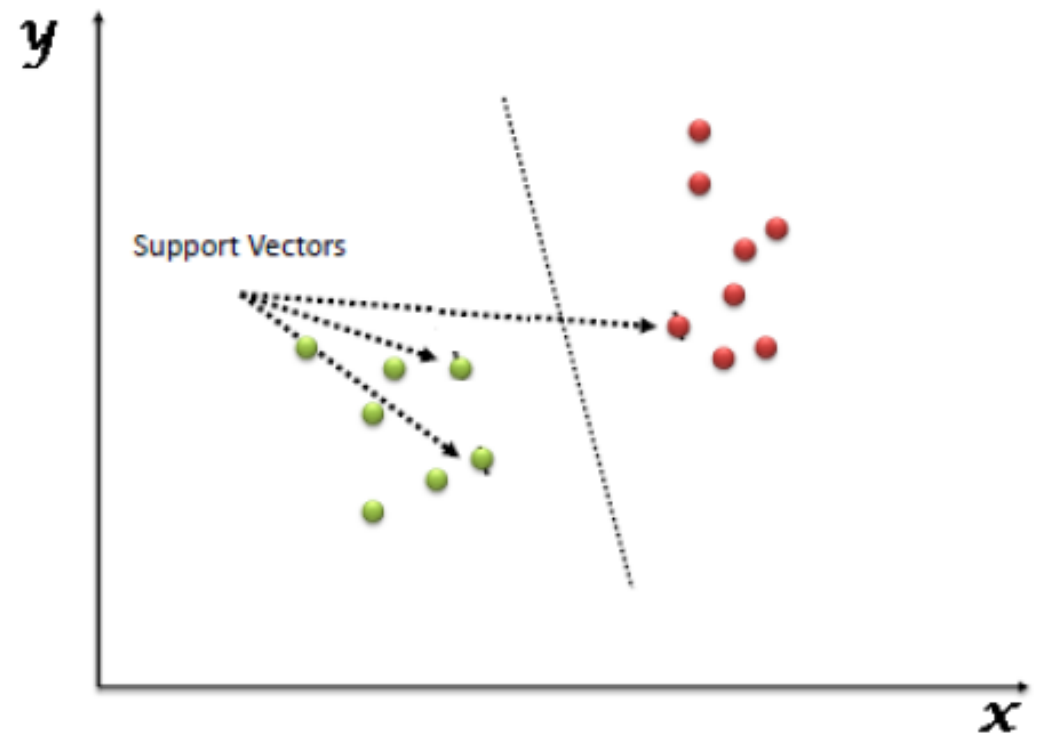

Fig. 5 Support Vector Machine

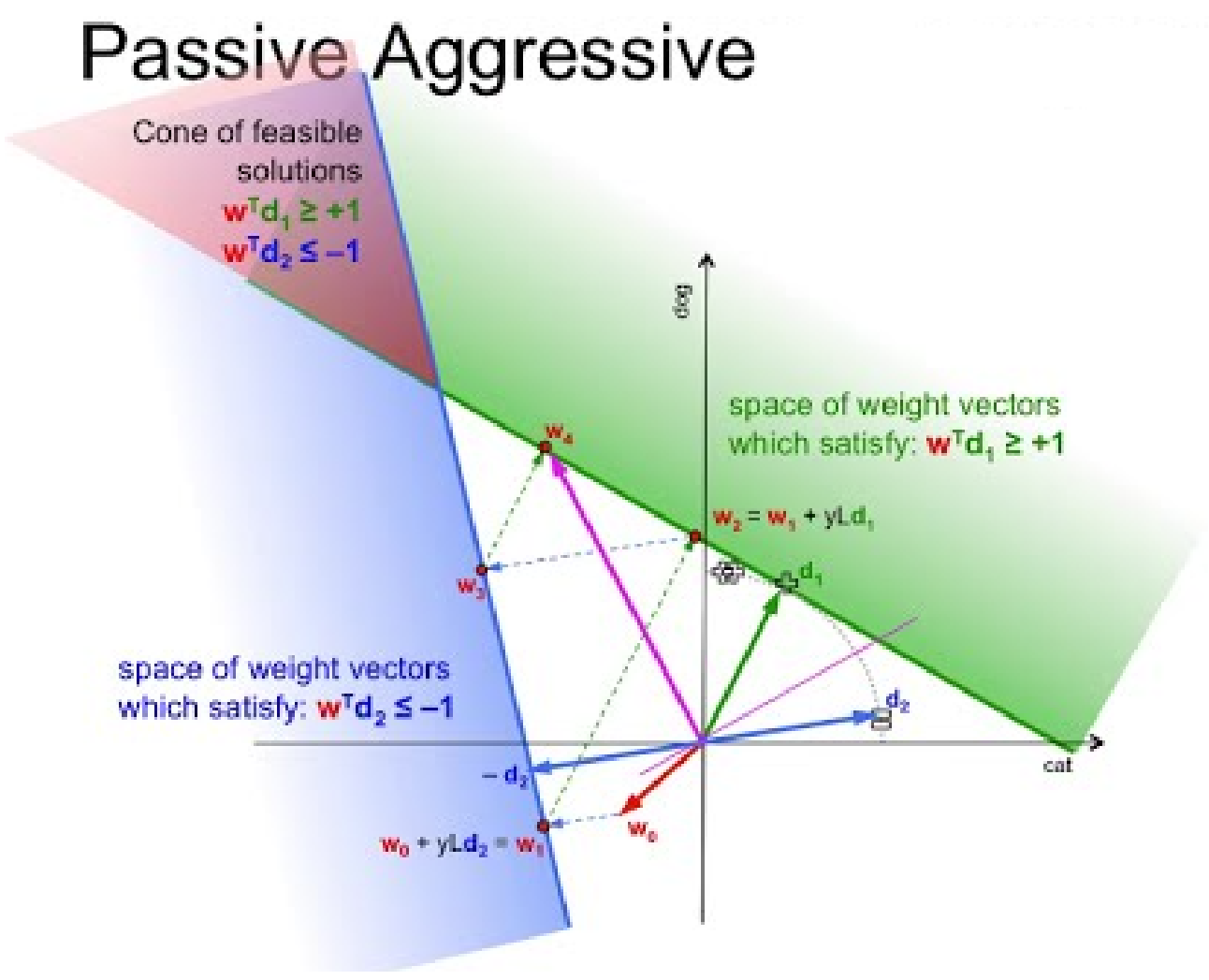

Fig. 6 Passive Aggressive Classifier

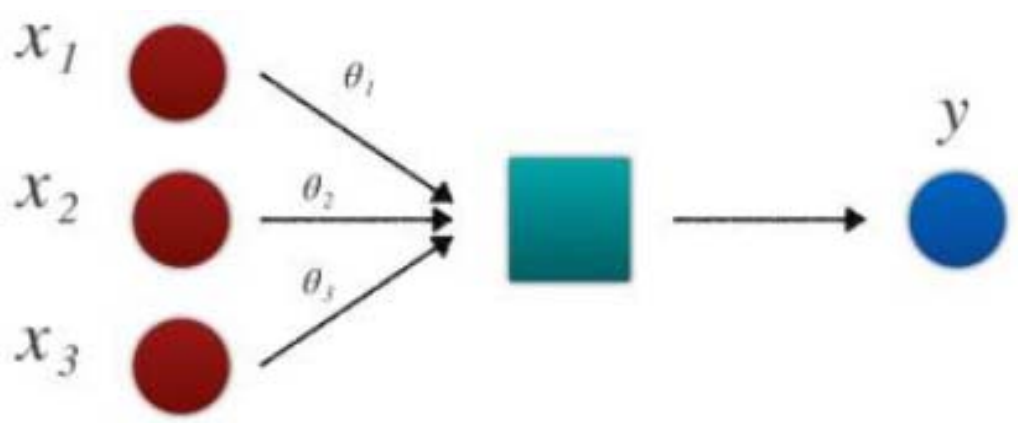

Fig. 7 Logistic Regression Model

## IV. Proposed Model

Our proposed model starts with the extraction phase and then we have four main steps. The first step is related to the NLP models where we measure the frequency of words and build the vocabulary of known words in fake news datasets. Next, fake news is detected using NB, SVM and PA classifiers. Finally, we test our models with several experiments and some other datasets and propose the final fake news detection model. Fig. 9 shows the flowchart of our model.

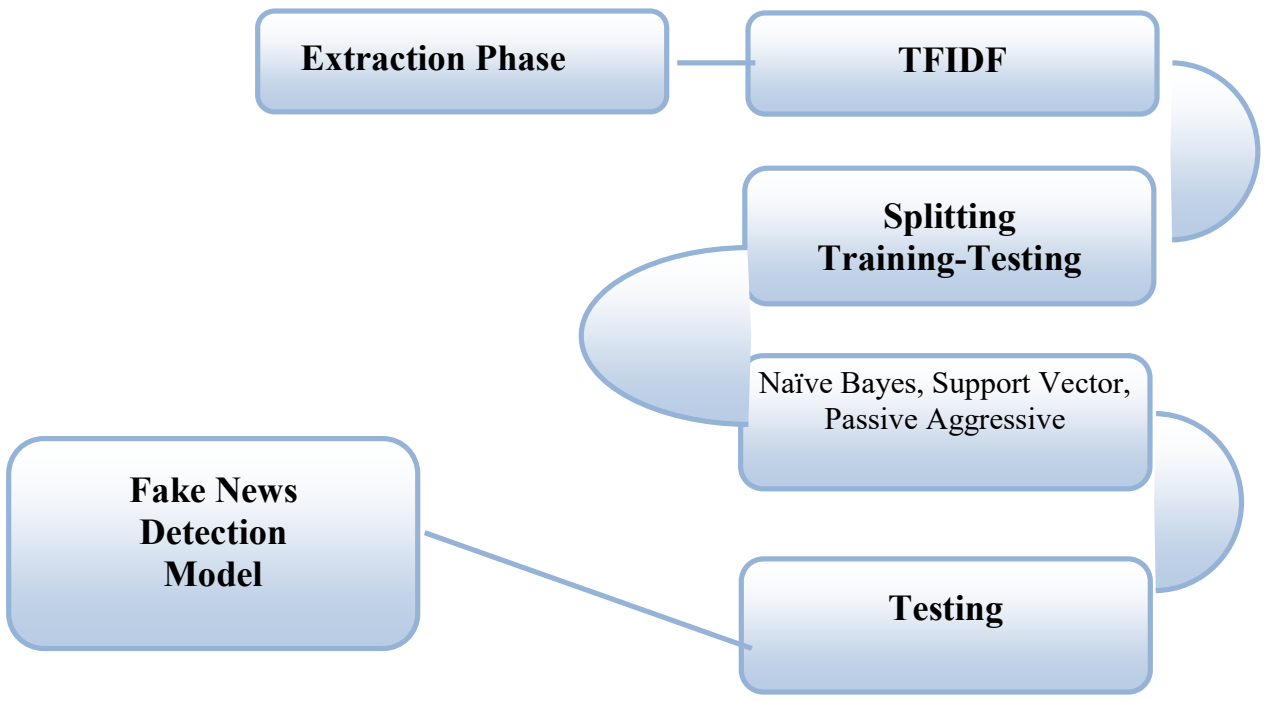

Fig. 8 Fake News Detection Model

## V. Preprocessing

The objective of this phase is to reduce the size of the data by removing irrelevant information that is not necessary for classification. Subsequently, for processing, the data were changed so that the first half of the data with the fake label set and the second half with a true label were not simply what would cause impartiality when applying the machine learning methods. One common task in NLP is tokenization that takes a text or set of texts and breaks it up into individual words. We converted words to their base form for better understanding [21]. Then we applied stemming that decreases the number of words on the bases of word type and class. Let us suppose we have three similar words in the dataset like running, ran and runner; it will be reduced and changed to the word, run. There are different stemming algorithms, but we used Porter due to its high accuracy rate. We used stop word removal as it removes common words used in articles, prepositions and conjunctions [24].

## VI. Machine Learning methods

Fake news is increasing every second without proper checks and balances, so there is a need for computational tools that can handle this problem. Machine learning algorithms like "CountVectorizer", "TFIDFVectorizer", naïve Bayes, Support Vector Machine, Passive Aggressive Classifier and NLP for the identification of false news in public data sets are proposed. This is purely a text-based classification problem but our actual goal is the combination of text-based classification with machine-based text transformation and then choosing which type of text is to be used, e.g. single news or the full body of the news. The overall data cleaning process is shown in Fig. 9.

### *A. NLP Models*

Irrelevant and redundant features in a dataset have a negative impact on the accuracy and performance of the classifier. So, in those cases, we perform feature reduction to reduce the text feature size that limited the words like "the", "and", "there", "when" and focus only on those words which appear a given number of times. This is done by using n-number of use words, lower casing and stop word removal since the sensitivity of the problem, which is increasing every second without check and balance, is understood [22]. It is essential to use machine learning algorithms like CountVectorizer and TF-IDF to speed up the task and improve performance.

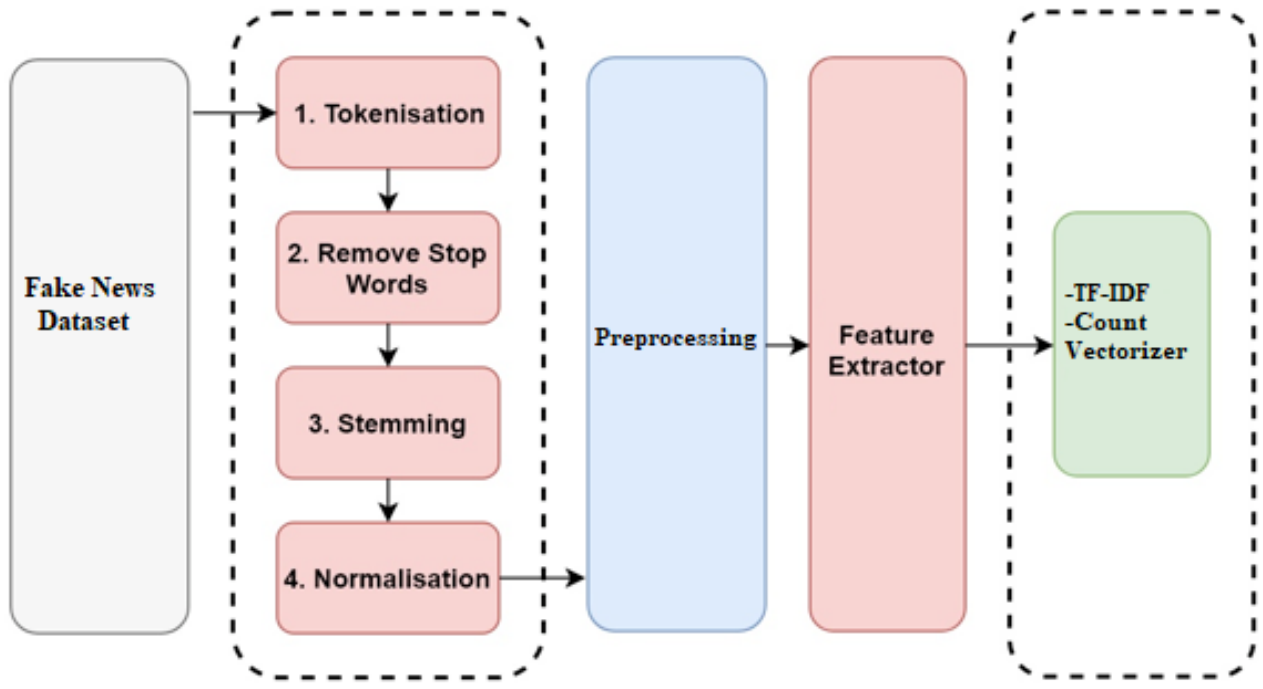

Fig. 9 Data cleaning steps starting from raw dataset to Machine learning models

### *B. Count Vectorization*

CountVectorizer provides a simple way to collect text documents and to help build the vocabulary of known distinctive words and also to encode new documents using that vocabulary [23]. Given a collection of text documents, S to Count Vectorizer and it will generate a sparse matrix of size A where m = total number of documents, n = total number of distinct words used in S. With the Count Vectorizer, we can produce a table for each word and occurrence of each class.

### *C. Term Frequency- Inverse Document Frequency*

To measure a term in documents over a dataset, we used the term frequency-inverted document frequency. A term's importance increases in the document which appears in the dataset and also the frequency of the words. So, with the help of this method, we can weigh the metric that is used for information retrieval [24]. TF-IDF for the word with respect to document d and corpus D is calculated as follows:

$$TF - IDF(w)_{d,D} = TF(w)_d \times IDF(w)_D. \qquad (1)$$

Let us suppose we have a document with 100 words and we need to calculate TF-IDF for the word "rumor." The word "rumor" occurs in the document 4 times; then we can calculate, TF = 4/100 = 0.04. Now, we need to calculate the IDF; let us assume that we have 200 documents, and "rumor" appears in 100 of them. Then, IDF(rumor) = 1+log (200/100) = 0.5, and TF-IDF(rumor) = 0.05 × 0.5 = 0.025.

## VII. Implementation

In the study carried out, NLP is used as a Python computational tool, which uses different libraries and platforms. We applied PANDAS (Python Data Analysis Library) which is an open-source library with BSD license that provides data structures and data analysis tools during classification.

We applied NLTK in the extraction and characterization phase. Numpy and Scipy libraries are applied for programming but our main program is run on Jupyter Notebook. Keeping in mind the training and testing data, we further attached test data with tokenization algorithms. The main objective is to develop a model based on the count vectorization and TF-IDF. Fake news detection is a binary classification task that the news is fake or not fake [16]. Classification is not completely correct in fake news detection [12] because classification methods are not specialized for fake news detection. So, keeping in mind that classification can separate fake text from non-fake, the goal is to develop a model that is specialized for fake news detection [28]. To develop a classification method that is specialized for fake news detection we need to identify relevant features before classification. We applied different features to extract optimal features in the text that help us for better text classification.

## VIII. Results & Discussion

We conducted several experiments with different feature set combinations as discussed in Section III *C* and the model selection in Section III *B*. Our proposed combination works well and obtains performance above the baseline 0.50. The best performing classifier is PA when we check the performance through accuracy and precision. However, somehow in the recall it reduced. Table II shows the performance of our proposed classifiers.

TABLE II
Observed Results

| Classifier | Accuracy | Precision | Recall |
|---|---|---|---|
| Naïve Bayes | 0.85% | 0.89% | 0.87% |
| **Passive Aggressive** | **0.93%** | 0.92% | 0.89% |
| Support Vector Machine | 0.84% | 0.82% | 0.87% |

Fig. 10 displays the performance metrics of all the classifiers. The next section will describe the results when we compare the proposed combination with other datasets and different classifiers but in the same domain.

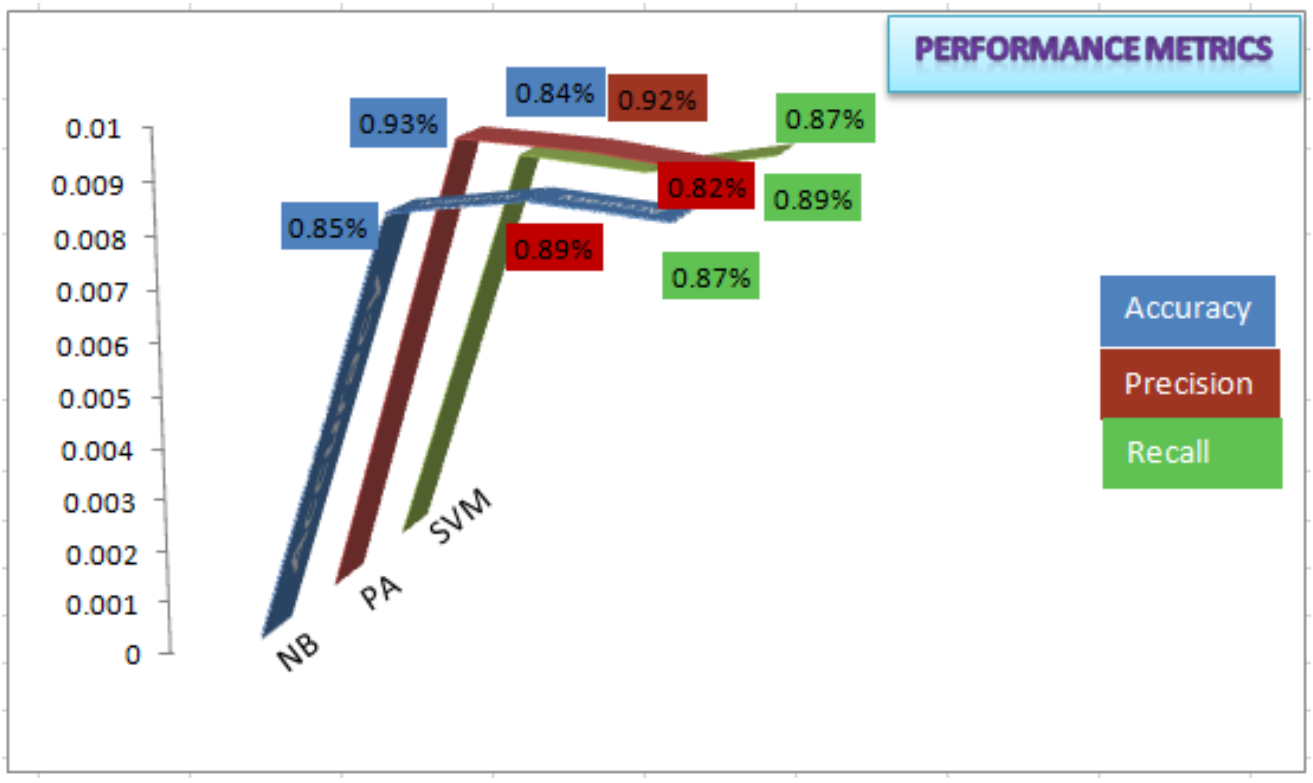

Fig. 10 Performance Metrics

### *A. Performance Comparison*

We compare our results with the same model but different datasets and different features, as highlighted in Table III. It is observed that the proposed models perform well and achieved the highest accuracy up to 93% with Passive Aggressive, 85% with naïve Bayes and 84% with SVM. Ott et al. applied SVM with features LIWC+ Bigrams and achieved an accuracy level of up to 89%. Similarly, when they changed the Stylometric features, it achieved 84% accuracy [17]. On the other side, Horrne and Adali achieved 71% accuracy when they applied text-based features [31]. The results show that the proposed combination improves the existing performance in some categories. For further analysis, we applied different combinations to check the accuracy of the proposed model with other models. Accuracy comparison of Passive Aggressive and Support Vector Machine (a), Passive Aggressive and Logistic Regression (b), Passive Aggressive and Support Vector Machine (c) with a different dataset, Passive Aggressive and Naïve Bayes (d), Support Vector Machine and Naïve Bayes (e), Naïve Bayes and Support Vector Machine (f), Support Vector Machine and Logistic Regression (g) and Support Vector Machine and Naïve Bayes (h) can be seen in Fig. 11. We further investigated and compared our results with [2] when they applied a combination of CFG and n-gram accuracy in deception detection where they achieved 85%-91% accuracy. Still, our presented results are better in the context of fake news detection and our proposed classifiers achieved maximum accuracy level.

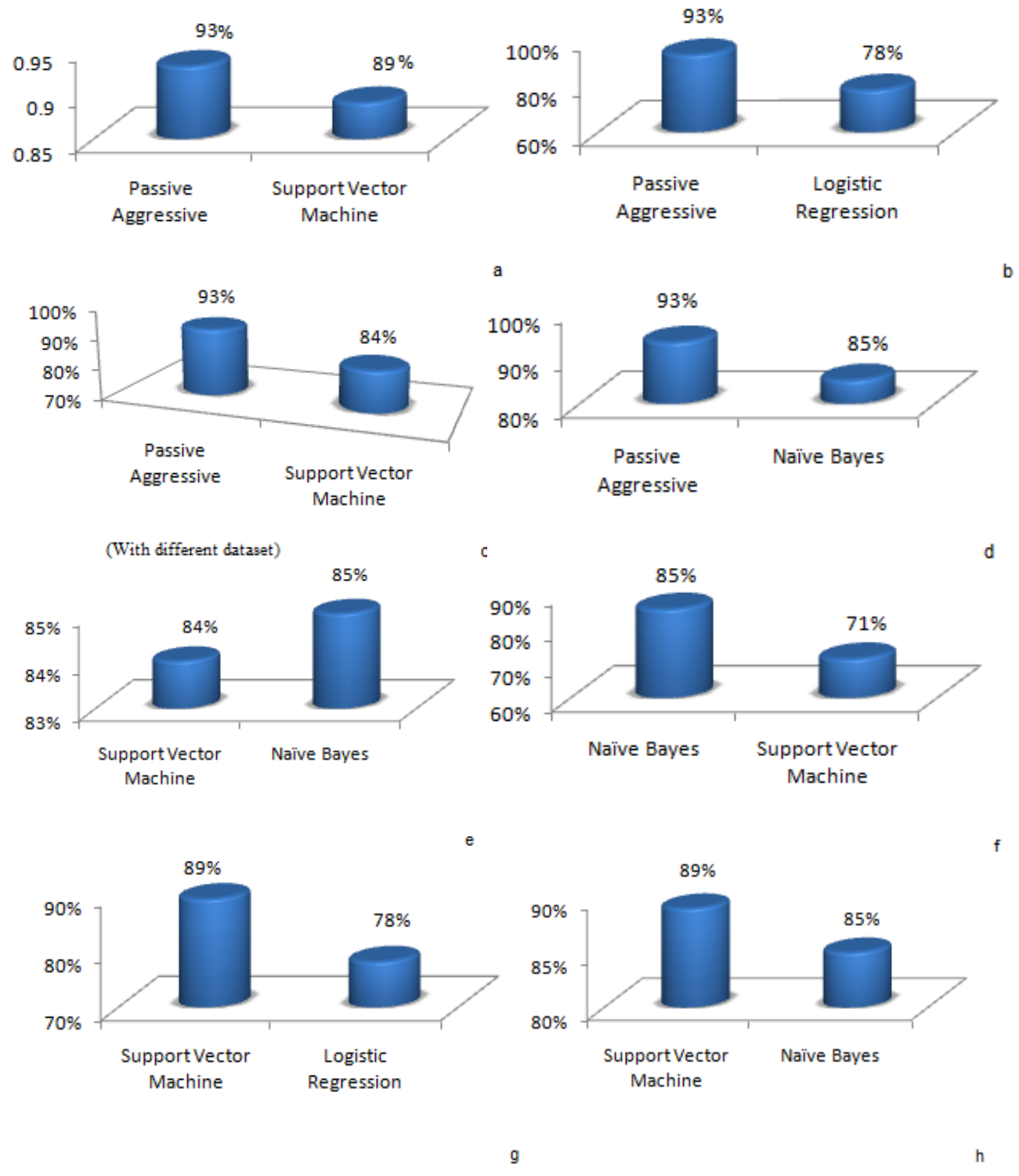

Fig. 11 Accuracy Comparison with different Algorithms

Figs. 11 (a)-(h) show the results of the classification for the PA, SVM and NB classifiers. The values are the maximum accuracy level achieved by the classifier after combining it with others. For further understanding of the results, we changed the classifier and fake news dataset proposed by others. These experiments highlighted some important features that we still want to investigate further.

TABLE III
PERFORMANCE COMPARISON WITH EXISTING APPROACHES

| Dataset | Classifier | Features | Performance Metrics | Score |
|---|---|---|---|---|
| Fake News Dataset | Passive Aggressive | News Articles | Accuracy | **0.93%** |
| Dataset (Ott et al. dataset 2017) [17] | Support Vector Machine | LIWC+ Bigrams | Accuracy | 0.89% |
| Reviews Amazon website (Jindal N et al. 2008) [32] | Logistic Regression | Review and reviewer feature | Accuracy | 0.78% |
| Dataset (Ott et al. dataset 2011) [17] | Support Vector Machine | Stylometric features | F-measure | 0.84% |
| Fake news data set | Naïve Bayes | News Articles | Accuracy | **0.85%** |
| Fake news data set | Support Vector Machine | News Articles | Accuracy | **0.84%** |
| Random new articles (Horne and Adali's news dataset 2019) [31] | Support Vector Machine | Text-based features | Accuracy | 0.71% |

## IX. CONCLUSION

The results suggested that the approach is highly favorable since this application helps in classifying fake news and identifying key features that can be used for fake news detection. Our proposed technique suggests that to differentiate fake and non-fake news articles, it is worthwhile to look at machine learning methods. The developed system with accuracy up to 93% proves the importance of the combination; next, we need to look into other methods for fake news detection except for simple text classification. The producers of fake news are using different techniques to hide their identity, so they can easily mislead readers. As we are aware that every single news has different characteristics so there is a need for a system that can check the content of the news in depth. Our future work includes building an automated fact-checking system that combines data and knowledge to help non-experts and checks the content of the news thoroughly after comparing it with known facts. We want to look into the issue of fake news from different angles like known facts, source, topic, associated URLs, geographical location, year of publication, and credibility of the source for a better understanding of the problem. The open issues and challenges are also presented in this paper with potential research tasks that can facilitate further development in fake news research.